# Micro-expression Recognition Based on Dual-branch Feature Extraction and Fusion

Mingjie Zhang#
*School of Petroleum*
*China University of Petroleum*
*(Beijing) Karamay Campus*
Karamay, China
2023591304@cupk.edu.cn

Bo Li#
*Artificial Intelligence Research Institute*
*China Tower Corporation Limited*
Beijing, China
libo88872@chinatowercom.cn

Wanting Liu#
*Low Carbon Business Department*
*South Electric Power Science and Technology Development Co., Ltd.*
Guangzhou, China
2407463533@qq.com

Hongyan Cui#*
*School of Information and Communication Engineering*
*Beijing University of Posts and Telecommunications*
Bejing, China
cuihy@bupt.edu.cn

Yue Li#
*School of Petroleum*
*China University of Petroleum*
*(Beijing) Karamay Campus*
Karamay, China
2022015399@st.cupk.edu.cn

Qingwen Li#
*Artificial Intelligence Research Institute*
*China Tower Corporation Limited*
Beijing, China
liqw3@chinatowercom.cn

Hong Li#
*Artificial Intelligence Research Institute*
*China Tower Corporation Limited*
Beijing, China
lihong@chinatowercom.cn

Ge Gao#
*School of Petroleum*
*China University of Petroleum*
*(Beijing) Karamay Campus*
Karamay, China
2609565842@qq.com

*Abstract*—Micro-expressions, characterized by transience and subtlety, pose challenges to existing optical flow-based recognition methods. To address this, this paper proposes a dual-branch micro-expression feature extraction network integrated with parallel attention. Key contributions include: 1) a residual network designed to alleviate gradient vanishing and network degradation; 2) an Inception network constructed to enhance model representation and suppress interference from irrelevant regions; 3) an adaptive feature fusion module developed to integrate dual-branch features. Experiments on the CASME II dataset demonstrate that the proposed method achieves 74.67% accuracy, outperforming LBP-TOP (by 11.26%), MSMMT (by 3.36%), and other comparative methods.

## I. Introduction

Micro-expressions reflect authentic subconscious feelings, serving as effective behavioral cues in criminal investigation, marketing, etc., and are significant for mental health and professional communication. Deep learning outperforms traditional methods in capturing deep semantic information for micro-expression recognition but faces challenges like insufficient training data and difficulty in fine-grained feature extraction.

To solve these, this paper proposes a dual-branch feature fusion framework: global and local facial key points (identified via CASME II action unit annotations) are input into ResNet and Inception modules respectively. Attention weights of global/local features are calculated and embedded into the fusion module to enhance fine-grained feature capture.

Main contributions: (1) A dual-branch micro-expression recognition model is proposed, integrating ResNet/Inception with convolutional block attention mechanism (CBAM) to focus on salient regions; (2) A CBAM-based feature fusion module is designed to address the lack of effective attention in dual-branch models; (3) Experiments on CASME II verify the method's superiority, achieving 74.67% accuracy which outperforms LBP-TOP and state-of-the-art methods.

## II. Related Work

Microexpression research has evolved from psychological foundations to computer vision-based automatic recognition. Early work established the Facial Action Coding System (FACS) proposed by Ekman, providing the theoretical basis for subsequent technical research.

Current recognition methods fall into three categories: **handcrafted feature-based approaches** (e.g., LBP and optical flow techniques [1]), which are computationally expensive and noise-sensitive; **deep learning-based methods** (e.g., DTSCNN [2], OFF-ApexNet [3], and related networks [4][5]) that have become mainstream due to their strong representational capacity; and **transfer learning-based techniques** (e.g., self-attention-enhanced transfer [6], domain adaptation [7], MMD-based transfer networks [8]) that mitigate data scarcity by leveraging macro-expression datasets.

Additionally, Inception architectures improve multi-scale feature extraction, while attention mechanisms (e.g., SE, SAM, CBAM) enhance discrimination by focusing on subtle facial movements. However, existing methods often rely on full video sequences, leading to information redundancy and computational complexity. To address the transient, local, and low-intensity nature of microexpressions, this study focuses on onset-apex frames, proposes a dual-branch global-local feature extraction framework, and integrates parallel attention with feature fusion to improve recognition performance.

## III. METHOD

### A. Problem Description

This paper proposes a dual-branch feature fusion framework for micro-expression recognition. Based on the alignment between visualized facial motion and annotated Action Units (AUs) in the CASME II dataset, facial key regions corresponding to global and local features are identified and separately fed into two attention modules built upon the ResNet and Inception architectures for adaptive feature weighting. The resulting global feature $F_G$ and local feature $F_L$ are then integrated and processed through a convolutional block attention-based feature fusion module, which dynamically adjusts the model's attention to different channels and spatial positions during training.

As illustrated in Figure 1, the proposed network consists of three core components: a global branch (ResNet), a local branch (Inception), and a convolutional block attention-based feature fusion module (CAFFM). The input micro-expression image F is processed in parallel through the two branches to extract $F_G$ and $F_L$, which are then fused and passed through three consecutive CBAM modules. After concatenation, ReLU activation, and residual addition, the features are further refined by two additional CBAM modules, followed by activation and max-pooling operations to accomplish the recognition task.

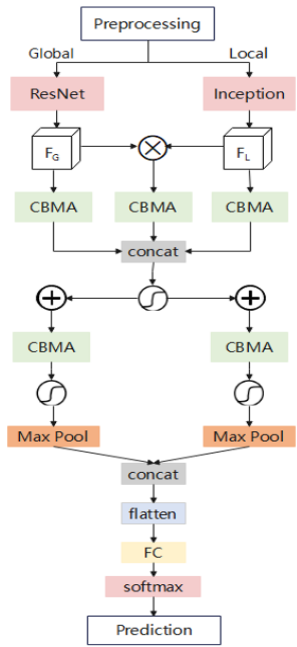

FIGURE 1. NETWORK ARCHITECTURE

### B. Dual-Branch Feature Extractio

For global feature extraction, to avoid problems such as gradient vanishing and degradation during the training process, a backbone network for feature extraction based on the ResNet architecture is introduced. By stacking residual blocks, skip connections are incorporated into the network, enabling it to learn deeper-level global features. The two main types of residual blocks in ResNet are shown in **Figure 2**.

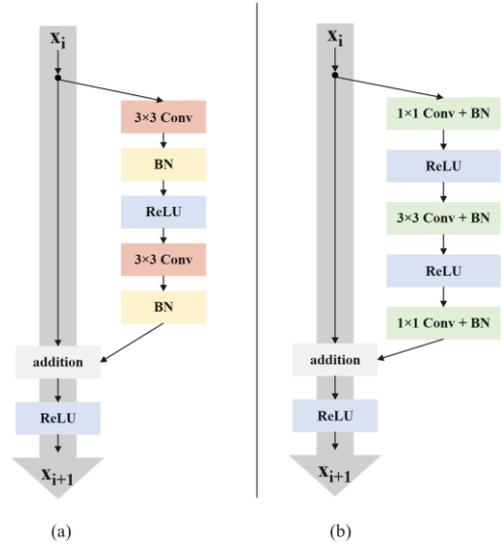

FIGURE 2. TWO TYPES OF CONVOLUTIONAL RESIDUAL BLOCKS IN RESNET

(a) BASIC BLOCK; (b) BOTTLENECK

Specifically, the residual block in **Figure 2(a)** is generally referred to as the Basic Block, which consists of two 3×3 convolutional layers. The residual block in **Figure 2(b)** is commonly known as the BottleNeck, which successively employs two 1 × 1 convolutional layers to achieve dimension expansion and dimension reduction of the feature maps. The primary purpose of this design is to address the issue of prolonged model training time caused by excessive parameters when a large number of residual blocks are stacked. In addition, the identity mapping of the skip connections can help the network learn more meaningful feature representations, thereby enhancing the representational capacity and generalization performance of the network.

## IV. EXPERIMENT

### A. Data Set

A total of 255 multi-frame micro-expression samples from the CASME II dataset were annotated to determine their corresponding Action Units (AUs). Given the extremely limited samples of Fear and Sadness and the fact that their AUs are fully covered by the "Others" category, these two categories were merged into the broader "Others" class. Based on the distribution of AUs across facial regions, facial feature extraction was divided into five areas: ocular and brow region (AU1, AU2, AU4, AU5, AU7), oral region (AU10, AU12, AU14, AU15, AU16, AU25, AU26), mandibular region (AU17), cheek region (AU6), and nasal region (AU9, AU38).

For accurate facial region localization and extraction, a four-dimensional Blob object was first constructed using OpenCV as the neural network input. A pre-trained DNN model from TensorFlow was loaded to perform forward propagation, yielding face detection results and confidence values (see Figure 3), based on which the facial region was cropped.

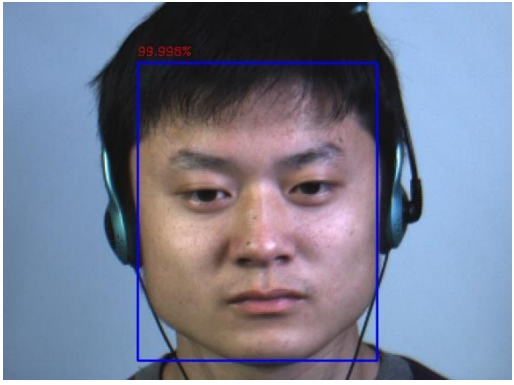

FIGURE 3. DNN FACE DETECTION AND CROPPING ALIGNMENT

Subsequently, facial images were standardized and resized to 231×282 pixels to meet the input requirements of both the ResNet model and the Inception structure.

### B. Experimental Setup

**Experimental parameters:** The batch size was set to 64 and the number of epochs was set to 500. A decay strategy was adopted to adjust the learning rate, with the initial learning rate set to 0.001. The decay strategy was step decay, which was performed once every 100 epochs with the gamma value set to 0.9. The Adam optimizer was selected for model training.

### C. Ablation Experiments

Ablation experiments were conducted to compare the performance differences of ResNet18, ResNet34 and ResNet12 on the micro-expression recognition task. The experiments were carried out on the CASME II dataset with global feature images as the input data. Accuracy, UF1, and UAR were adopted as the evaluation metrics. The Accuracy, UF1 and UAR values of the three models are presented in **Table 5** and **Figure 4**.

TABLE 5. RESNET WITH DIFFERENT NUMBER OF LAYERS

| ResNet_X | Evaluation Metrics | | |
|---|---|---|---|
| | *Accuracy* | *UF1* | *UAR* |
| ResNet_12 | 0.7577 | 0.7610 | 0.7643 |
| ResNet_18 | 0.7434 | 0.7282 | 0.7137 |
| ResNet_34 | 0.7515 | 0.7398 | 0.7284 |

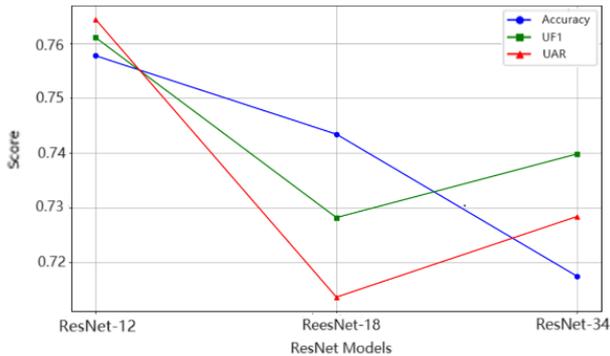

FIGURE 4. RESNET WITH DIFFERENT NUMBER OF LAYERS

It can be seen from the experimental results that, compared with the corresponding metrics of ResNet12 on the CASME II dataset, the Accuracy, UF1 and UAR of ResNet18 decreased by 0.0143, 0.0328 and 0.0506, respectively. In contrast, the Accuracy, UF1 and UAR of ResNet34 decreased by 0.0402, 0.0212 and 0.0359, respectively, compared with those of ResNet12 on the same dataset. The experimental results demonstrate that a tendency is observed on the CASME II dataset where the performance of the ResNet model decreases instead of increasing with the increase in the number of network layers. This phenomenon may be attributed to the small sample size of the micro-expression dataset itself: when the model is excessively deep, the dataset scale is mismatched with the model complexity, which consequently leads to overfitting. Therefore, ResNet12 was selected as the backbone network for subsequent research in this paper.

### D. Comparative Experiments

TABLE 7. ABLATION EXPERIMENTS ON DIFFERENT FEATURE MODULES

| Method | Acc/% | Parameters/M | Complexity/G | FPS |
|---|---|---|---|---|
| GFEM | 66.31 | 2.59 | 1.778 | 111.5 |
| LTFEM | 62.69 | 6.26 | 3.287 | 106.3 |
| DBFEM | 71.16 | 8.13 | 4.927 | 104.5 |
| DBFEM+CAFFM | 74.67 | 8.46 | 5.165 | 97.3 |
| DBFEM+$CAFFM_L$ | 73.52 | 8.49 | 5.14 | 97.3 |
| DBFEM+$CAFFM_G$ | 72.23 | 8.5 | 5.151 | 97.5 |

As shown in **Figure 6**, the confusion matrix of the proposed model is presented on the CASME II dataset, which explains the reason for the relatively low recognition rate of the model on this dataset. For instance, the facial regions corresponding to Surprise and Repression are both associated with the action units of mouth corner movements. The similarity of these AU activation regions is prone to cause misclassification, thus leading to a low recognition rate.

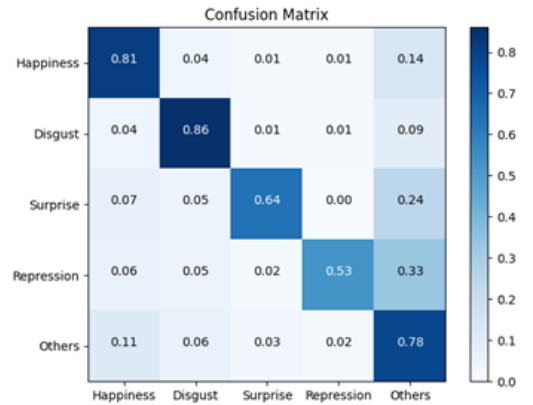

FIGURE 6. CONFUSION MATRIX ON CASME II DATASETS

To investigate the accuracy of the proposed scheme for micro-expression recognition, the model was trained and tested on the CASME II dataset, and the test results were compared with those of various state-of-the-art methods, with the experimental results shown in **Table 8**. The proposed method achieved an accuracy of 74.67% on the CASME II dataset, which is 3.36%, 3.99% and 0.88% higher than those of the recently proposed methods (MSMMT, Later and SLSTT-Mean), respectively. The AMAN method, which adopts micro-expression magnification to enhance the intensity of micro-expressions, achieved an accuracy of 75.4%. The reason why the accuracy of the proposed method is slightly lower than that of AMAN may be attributed to the fact that the proposed method uses raw data images without any magnification processing. Nevertheless, most samples of each expression category can be accurately identified by the proposed method, which reflects the effectiveness and reliability of the proposed method.

TABLE 8. CASME II DATASET RECOGNITION ACCURACY

| Method | Accuracy | Precision | Recall |
| --- | --- | --- | --- |
| LBP-TOP | 63.41 | 0.628 | 0.635 |
| DISTLBP-IIP | 64.78 | 0.642 | 0.649 |
| 3D-CNNs(with | 65.9 | 0.653 | 0.66 |
| SSSN[10] | 71.19 | 0.706 | 0.713 |
| AKMNet[11] | 67.06 | 0.665 | 0.672 |
| KFC-MER[12] | 72.76 | 0.721 | 0.728 |
| FR[13] | 62.85 | 0.622 | 0.629 |
| Knowledge | 72.61 | 0.72 | 0.727 |
| SLSTT-Mean[15] | 73.79 | 0.732 | 0.739 |
| Later[16] | 70.68 | 0.701 | 0.708 |
| KTGSL[17] | 72.58 | 0.719 | 0.726 |
| AMAN[18] | 75.4 | 0.748 | 0.755 |
| MSMMT[19] | 71.31 | 0.707 | 0.714 |
| **OURS** | **74.67** | 0.741 | 0.748 |

## V. CONCLUSION

This paper proposes a micro-expression feature extraction method combining a dual-branch network of ResNet12 and Inception v3, along with a feature fusion module based on the convolutional block attention mechanism. By simultaneously leveraging global and local features, the model learns subtle motion characteristics while accounting for the local attributes of micro-expressions. The backbone network embedded with CAFFM is constructed by stacking CBAM blocks, allowing adaptive weight adjustment across channels and spatial positions. Ablation experiments confirm the effectiveness of the proposed method and its optimization strategies, with the model generally meeting real-time requirements. Comparative experiments on the CASME II dataset further demonstrate its superiority.

Future work will focus on: constructing large-scale, high-quality micro-expression datasets to facilitate large-sample research; developing models with stronger generalization ability for cross-dataset recognition; and designing algorithms that integrate both micro-expression detection and recognition for practical deployment.

## ACKNOWLEDGMENTS

This work was supported by National Natural Science Foundation of China (62171049), China Tower Corporation Limited IT System 2023 Package Software Project - AI Algorithm and Services (23M01ZBZB011000017), and China University of Petroleum (Beijing) Karamay Campus introduction of talents and launch of scientific research projects(XQZX20240010). All authors are contributed equally to this work.

## REFERENCES

[1] Davison A K, Merghani W, Yap M H. Objective classes for micro-facial expression recognition[J]. Journal of imaging, 2018, 4(10): 119.

[2] Khor H Q, See J, Phan R, et al. Enriched Long-term Recurrent Convolutional Network for Facial Micro-Expression Recognition[C].2018 13th IEEE International Conference on Automatic Face & Gesture Recognition, Xi'an, China&Washington, USA,2018:667-674.

[3] Gan Y S, Liong S T, Yau W C, et al. OFF-ApexNet on micro-expression recognition system[J]. Signal Processing: Image Communication, 2019, 74: 129-139.

[4] Verma M, Vipparthi S K, Singh G, et al. LEARNet: Dynamic imaging network for micro expression recognition[J]. IEEE Transactions on Image Processing, 2019, 29: 1618-1627.

[5] Cai L Q, Li H, Dong W, et al. Micro-expression recognition using 3D DenseNet fused Squeeze-and-Excitation Networks[J]. Applied Soft Computing, 2022, 119(1): 1-12.

[6] Indolia S, Nigam S, Singh R. Integration of 'Transfer Learning and Self-Attention for Spontaneous Micro-Expression Recognition[C]. 2022Seventh International Conference on Parallel，Distributed and GridComputing(PDGC), JiNan,China,2022:325-330.

[7] Yan K, Zheng W, Cui Z, et al.Unsupervised facial expression recognition using domain adaptation based dictionary learning approach[J]. Neurocomputing, 2018, 319: 84-91.

[8] Xia W, Zheng W, Zong Y, et al. Motion attention deep transfer network for cross-database micro-expression recognition[C]. Pattern Recognition. ICPR International Workshops and Challenges: Virtual Event,Tokyo,Japan, 2021: 679- 693.

[9] Zhi R, Xu H, Wan M, et al. Combining 3D convolutional neural networks with transfer learning by supervised pre-training for facial micro-expression recognition[J].IEICE Transactions on Information and Systems, 2019, 102(5): 1054-1064.

[10] Khor H Q, See J, Liong S T, et al. Dual-stream shallow networks for facial micro-expression recognition[C]//2019 IEEE International Conference on Image Processing (ICIP), 2019: 36-40.

[11] Peng M, Wang C, Gao Y, et al. Recognizing micro-expression in video clip with adaptive key-frame mining[J].

[12] Su Y, Zhang J, Liu J, et al. Key facial components guided micro-expression recognition based on first & second-order motion[C]//2021 IEEE International Conference on Multimedia and Expo (ICME). IEEE, 2021: 1-6.

[13] Zhou L, Mao Q, Huang X, et al. Feature refinement: An expression-specific feature learning and fusion method for micro-expression recognition[J].Pattern Recognition, 2022, 122: 108275.

[14] Sun B, Cao S, Li D, et al. Dynamic micro-expression recognition using knowledge distillation[J].IEEE Transactions on Affective Computing, 2020, 13(2): 1037-1043.

[15] Zhang L, Hong X, Arandjelović O, et al. Short and long range relation based spatio-temporal transformer for micro-expression recognition[J]. IEEE Transactions on Affective Computing, 2022, 13(4): 1973-1985.

[16] Hong J, Lee C, Jung H. Late fusion-based video transformer for facial micro-expression recognition[J]. Applied Sciences, 2022, 12(3): 1169.

[17] Wei J, Lu G, Yan J, et al. Learning two groups of discriminative features for micro-expression recognition[J]. Neurocomputing, 2022, 479: 22-36.

[18] Wei M, Zheng W, Zong Y, et al. A novel micro-expression recognition approach using attention-based magnification-adaptive networks[C] //ICASSP 2022-2022 IEEE International Conference on Acoustics, Speech and Signal Processing (ICASSP). IEEE, 2022: 2420-2424.

[19] Wang F, Li J, Qi C, et al. Multi-scale multi-modal micro-expression recognition algorithm based on transformer[J].2023,arXiv preprint arXiv:2301.02969:1-20.